\documentclass[final]{cvpr}

\usepackage{graphicx}
\usepackage{amsmath}
\usepackage{amssymb}
\usepackage{amsfonts}
\usepackage{bm}
\usepackage{xcolor}
\usepackage[]{algorithm2e}
\makeatletter
\renewcommand{\@algocf@capt@plain}{above}
\makeatother

\usepackage{siunitx}
\usepackage[pagebackref=true,breaklinks=true,colorlinks,bookmarks=false]{hyperref}

\def\cvprPaperID{ID11} 
\def\confYear{CVPR 2021}

\begin{document}
\title{Out-of-distribution Detection and Generation using Soft Brownian Offset Sampling and Autoencoders}

\author{Felix Möller, Diego Botache, Denis Huseljic, Florian Heidecker, Maarten Bieshaar,\\ and Bernhard Sick\\         Intelligent Embedded Systems, University of Kassel, Germany\\
        \{fm $\mid$ diego.botache $\mid$ dhuseljic $\mid$ florian.heidecker $\mid$ mbieshaar $\mid$ bsick\}@uni-kassel.de\\
        www.ies.uni-kassel.de}

\newcommand{\felix}[1]{{\color{violet}#1}}
\newcommand{\denis}[1]{{\color{blue}#1}}
\newcommand{\diego}[1]{{\color{magenta}#1}}

\maketitle

\begin{abstract}
Deep neural networks often suffer from overconfidence which can be partly remedied by improved out-of-distribution detection. For this purpose, we propose a novel approach that allows for the generation of out-of-distribution datasets based on a given in-distribution dataset. This new dataset can then be used to improve out-of-distribution detection for the given dataset and machine learning task at hand. The samples in this dataset are with respect to the feature space close to the in-distribution dataset and therefore realistic and plausible. Hence, this dataset can also be used to safeguard neural networks, i.e., to validate the generalization performance. Our approach first generates suitable representations of an in-distribution dataset using an autoencoder and then transforms them using our novel proposed Soft Brownian Offset method. After transformation, the decoder part of the autoencoder allows for the generation of these implicit out-of-distribution samples. This newly generated dataset then allows for mixing with other datasets and thus improved training of an out-of-distribution classifier, increasing its performance. Experimentally, we show that our approach is promising for time series using synthetic data. Using our new method, we also show in a quantitative case study that we can improve the out-of-distribution detection for the MNIST dataset. Finally, we provide another case study on the synthetic generation of out-of-distribution trajectories, which can be used to validate trajectory prediction algorithms for automated driving.
\end{abstract}


\vspace{-.7cm}
\section{Introduction}
\begin{figure}[t]
    \centerline{\includegraphics[width=0.9\linewidth]{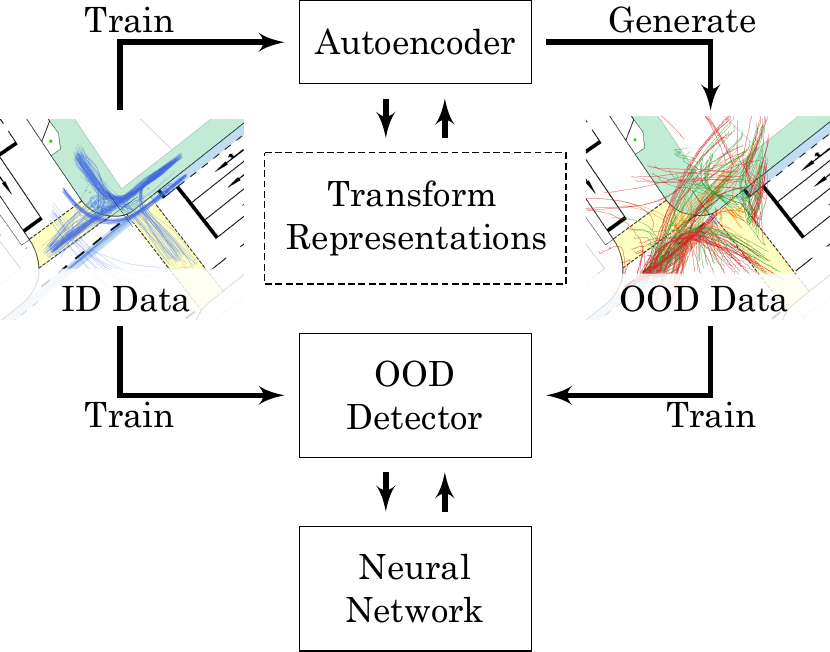}}
    \caption{Schematic overview of our approach to create OOD samples used for training OOD detectors or to directly validate an AI model. First, ID samples are used to train an Autoencoder (AE).
    Afterwards, we generate OOD samples using Soft Brownian Offset sampling on the latent  representation of the AE. The AE again decodes the OOD samples to form OOD samples in the original space. Finally, the novel OOD samples are used in conjunction with the original ID samples to train an OOD detector or to validate/ safeguard the AI model directly, i.e., neural model.}
\label{fig:overview}
\end{figure}

Artificial intelligence (AI) is the key technology for perception in automated driving. In particular, deep neural networks (DNN) are widely used in relevant tasks such as object detection~\cite{RHG+17,HHB+20} or trajectory prediction~\cite{BRZ+17,Bie21}. 
Safeguarding neural networks in terms of Safe AI~\cite{Hou+20} is of tremendous importance for safe automated driving.
Neural networks perform well when the distribution of the training and test data are sufficiently similar.
However, if they are too dissimilar, they can suffer from overconfidence~\cite{LPB17,AOS+16,GPS+17} which may even result in fatal events~\cite{NHTSA17}. 

A central role in safeguarding AI-function for perception in automated driving is being taken by the so-called corner cases, which are rare but mostly highly critical and therefore relevant cases~\cite{HHB+20,HBR+21}. The detection of such corner cases in machine learning (ML) is often referred to as out-of-distribution (OOD) samples~\cite{VGA+19,RLF+19,LLL+18} and is an essential building block for safeguarding AI-methods.
The AI methods can be validated and improved using these OOD samples. A complementary approach is the systematic artificial generation of corner cases and OOD samples, e.g., novel objects, critical (traffic) scenarios, or unusual trajectories of road users. 

This article presents a novel data-driven method for the artificial generation of OOD samples, which can be used to safeguard AI models and train improved AI-based OOD detectors. At its core, the method comprises a new algorithm, referred to as Soft Brownian Offset sampling, to create OOD samples at the tails of the data distribution used for training the AI model. Hence, the samples are OOD but still close to the actual in-distribution (ID) samples and therefore likely to be realistic and close to the AI model's intended operation domain.

Gaussian Hyperspheric Offset is a common baseline to create OOD samples by sampling along a normally distributed hypersphere around the ID-data.
Soft Brownian Offset Sampling is an algorithm based on the former method that allows to take an arbitrary point from a dataset and translate it to have a likely minimum distance to all other points from the dataset. The algorithm is no longer limited to sampling from a hypersphere around the ID data and is applied to the representations of an autoencoder (AE) trained to encode samples from a particular dataset.
A schematic of the proposed approach is given in Fig.~\ref{fig:overview}.
Using the AE's decoder's output for the transformed representations allows for generating additional datasets that are implicitly OOD. 
These datasets can then be used to improve an OOD detector's training or validate the AI model's functionality.

\subsection{Main Contributions}
The main contributions of this article can be summarized as follows:
\begin{itemize}
    \item We propose a novel algorithm, called Soft Browning Offset sampling, to create synthetic OOD samples (e.g., from the latent representation of an auto-encoder). We prove the applicability of the algorithm to different data modalities that are important for the perception of highly automated vehicles, i.e., images, time series, and trajectories. 
    \item In a case study with cyclists trajectories, we show that the novel OOD generation method creates unusual but still realistic trajectories to be used for validation of AI-based behavior prediction methods (cf.\ \cite{ZRK+19})
    \item We show that using synthetically generated OOD samples originating from the new algorithm improves the performance of state-of-the-art OOD Detectors.
    \item We include an easy-to-use Python implementation of the proposed OOD generation methods.\footnote{\url{https://pypi.org/project/sbo/}}
\end{itemize}

The remainder of this article is structured as follows: In Section~\ref{sec:related_work}, we review the current state of the art in the field of OOD-generation and detection. In Section~\ref{sec:methodology}, we detail the fundamentals of our approach and present our novel OOD sampling algorithm. Subsequently, we present experimental results in Section~\ref{sec:experiments} and in conclude our findings in Section~\ref{sec:conclusion}.

\section{Related Work} \label{sec:related_work}

\subsection{Out-of-distribution Detection}

In general, OOD detection can be described as the task of distinguishing between data that stems from one distribution and data that stems from another distribution that is sufficiently different. It can be formulated as a learning task in which we aim to separate samples originating of the in-distribution (ID) from those of the out-distribution (OOD)~\cite{CLW+20,Bel18}.

A neural-network-based OOD detection baseline is presented by Hendrycks and Gimpel \cite{HG+16}.
It does not require any retraining and utilizes the probabilities from softmax distribution to distinguish between in- and out-of-distribution samples. 
Another improved OOD detector is the ODIN detection, which uses temperature scaling on the trained network and small perturbations on inputs to separate ID and OOD samples based on the network's softmax score~\cite{LLS18}. Chen et al. propose the ALOE algorithm that improves the robustness of state-of-the-art OOD detectors\cite{CLW+20} by a novel robust training procedure incorporating both adversarially crafted ID and OOD samples.
Density-based approaches aim to build a probabilistic model of the data and then subsequently use this model for OOD detection, e.g.,~\cite{RLF+19}.
OOD detection is strongly related to tasks such as anomaly, or novelty detection~\cite{GS18} in which the goal is to detect unknown and potentially anomalous patterns.
Outlier Exposure~\cite{HMD+19} is a novel state-of-the-art deep anomaly detection, which uses a 
modified loss function to incorporate samples of an auxiliary dataset to better detect OOD samples.
\cite{LLL+18} propose a modification of loss functions and a novel training method to distinguish between OOD and ID data.
Prior Networks~\cite{Malinin2018, Malinin2019} use the aleatoric and epistemic uncertainty for OOD detection using a Dirichlet distribution. 
Another related method is proposed by Huseljic~et~al.~\cite{huseljic2020separation}. They utilize the properties of a Dirichlet-Categorical distribution and are able to measure and separate aleatoric and epistemic uncertainty. This is achieved by combining two objective functions -- the first optimizing on ID samples, the second on OOD samples -- into one by means of a convex combination. Furthermore, they suggest a naive technique to generate OOD samples by means of adapting the latent space of a generative adversarial network (GAN)~\cite{goodfellow_generative_2014}. Their model is then able to detect OOD samples while also allowing for a reliable estimation of the risk coming with a decision of the trained DNN.
Both~\cite{huseljic2020separation} and~\cite{LLL+18} require a separate set of (artificially generated) OOD samples to train OOD detectors.


\subsection{Out-of-distribution Generation}

In contrast to OOD detection, OOD generation is a relatively new field of research. 
Recently, the generation of training samples through deep generative models, e.g., GAN~\cite{goodfellow_generative_2014} or variational autoencoder (VAE)~\cite{kingma_auto-encoding_2014}, attracted attention in the OOD community. Lee et al.~\cite{Lee2017} noticed that samples generated at the tails of data distributions can be exploited to improve OOD detection. They use these samples to fine-tune the output of a DNN. Moreover, the authors propose a new GAN objective, which allows the generation of samples in low-density regions of training distributions. 
As shown by Vernekar et al.~\cite{VGA+19}, this generation procedure requires a DNN with already well-working estimation of predictive uncertainty. Furthermore,~\cite{VGA+19} show that their approach fails on a simple 3D-example indicating even more significant difficulties in higher dimensions.
In contrast, Vernekar et al.~utilize a conditional VAE~\cite{VGA+19} and define two types of OOD samples surrounding the latent encoding on a learned manifold.
However, their approach is limited by the dimensionality of input images due to insufficient generative capabilities of VAE and the need for a Jacobian matrix defined over the entire data set leading to high computational cost.
An alternative GAN-based approach for OOD sample generation is proposed by Sricharan and Srivastava and show that for effective OOD detection, the generated OOD samples should cover and be close to the low-density boundary of in-distribution~\cite{SS18}.
Catching up the idea of these approaches, we propose multiple novel geometric transformations that alleviate generative models to generate OOD samples that are similar to but also sufficiently different from the ID.

\section{Methodology} \label{sec:methodology}


This section introduces concepts and strategies that serve as a basis for the proposed approaches. The basis of our approach is a compact, relatively low-dimensional but yet, expressive representation of the data. Based on this representation, we present three methods to generate OOD samples. 

\subsection{Feature Representation using Autoencoders}

Many natural data sources show the property of presenting a low-dimensional, possibly noisy, manifold embedded within the higher dimensional observed data space~\cite{Bis06}. Approaches such as the principal component analysis or autoencoders try to capture this property. We use this low-dimensional embedding as the basis for our generative modeling and the OOD sampling.

Autoencoders, originally proposed in \cite{RM+87}, implement a dimensionality reduction with bottleneck layer (i.e., layer with significantly fewer neurons than in the input space).
After training, AEs can be separated into two parts: the layers up to the bottleneck can be used as an encoder, and the remaining layers are used as a decoder. The loss function of vanilla AE measures the reconstruction error between the input and decoded sample. The activations of the bottleneck layer's neurons (also referred to as latent variables) comprise a new compact feature (latent) space. They are the starting point of our OOD sampling strategy, i.e., we generate OOD samples in the latent space using the algorithms proposed in the following and subsequently use the decoder to transform the generated data back into the original feature space. 
There are many extensions to the original autoencoder, e.g., regularized, sparse, denoising, contractive, and variational autoencoder (VAE)~\cite{goodfellow_deep_2016}, which can all be combined with our OOD sampling strategies. Especially, the usage of VAE, which posses a relatively ``smooth'' latent space, is attractive as a basis for OOD sampling. Moreover, as it offers the direct capability to generate new samples, it provides a decent baseline for OOD sampling (cf.\ Gaussian Hyperspheric Offset Sampling).

At this point, we would like to emphasize that the approaches presented in this article are not necessarily dependent on a learned representation, e.g., from autoencoders, but can deal with generic low-dimensional feature representations as well.


\subsection{Gaussian Hyperspheric Offset}

The proposed method of Gaussian Hyperspheric Offset~(GHO) projects a point $\bm{x} \in \mathbb{R}^D$ onto the surface of a hypersphere with the most likely radius of $\mu$.
This radius is normally distributed by $\bm{n} \sim \mathcal{N}(0, \bm{I}_D)$ and its standard deviation scales with $\sigma$.
Uniform sampling on the surface of the hypersphere is achieved by scaling the normally distributed vector $\bm{s} \sim \mathcal{N}(0, \bm{I}_D)$ by the inverse of its length~\cite{Sim+15}:

\begin{equation}
\gamma(\mu, \sigma) = \mu \bm{s} |\bm{s}|^{-1} + \sigma \bm{n}
\end{equation}

\begin{figure}[tb]
    \centering

    \def\svgwidth{\columnwidth}
    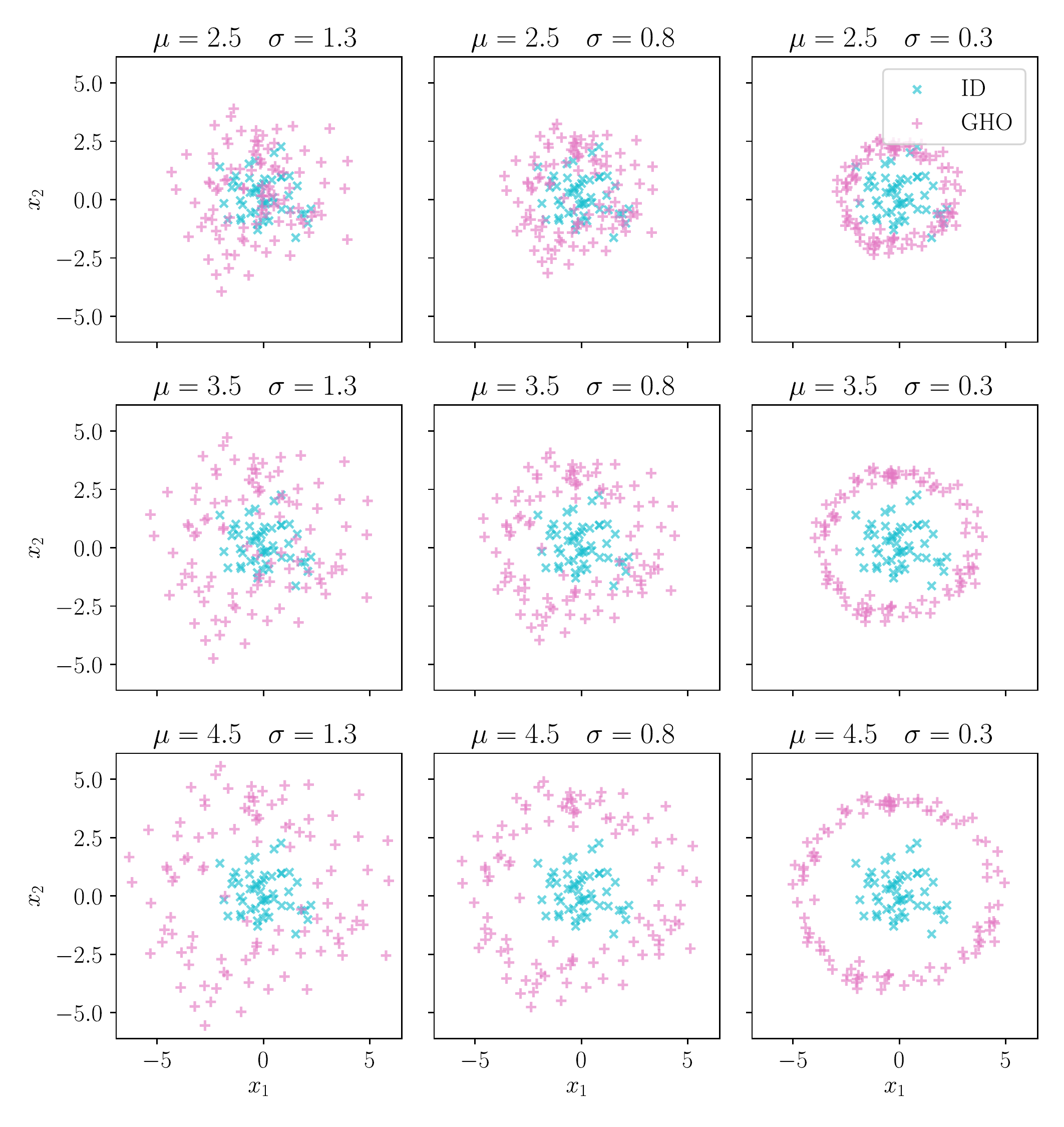
    
    \caption{Exemplary choice of parameters for GHO: ID dataset was sampled from $\bm{X} \sim \mathcal{N}(\bm{0}, \bm{I}_2)$ while OOD method's parameters are annotated. Varying both $\mu$ and $\sigma$ allows for precise control of OOD samples' behaviour.}
    \label{fig:gho}
\end{figure}

Fig.~\ref{fig:gho} shows the influence of the change of parameters of $\gamma$ in an exemplary fashion for $\mathbb{R}^2$.
One shortcoming of this method is its assumption to work on normally distributed data. 
While this may hold for specific applications our goal was to weaken these assumptions and propose a more general framework.

\begin{figure}[tb]
    \centering

    \def\svgwidth{\columnwidth}
    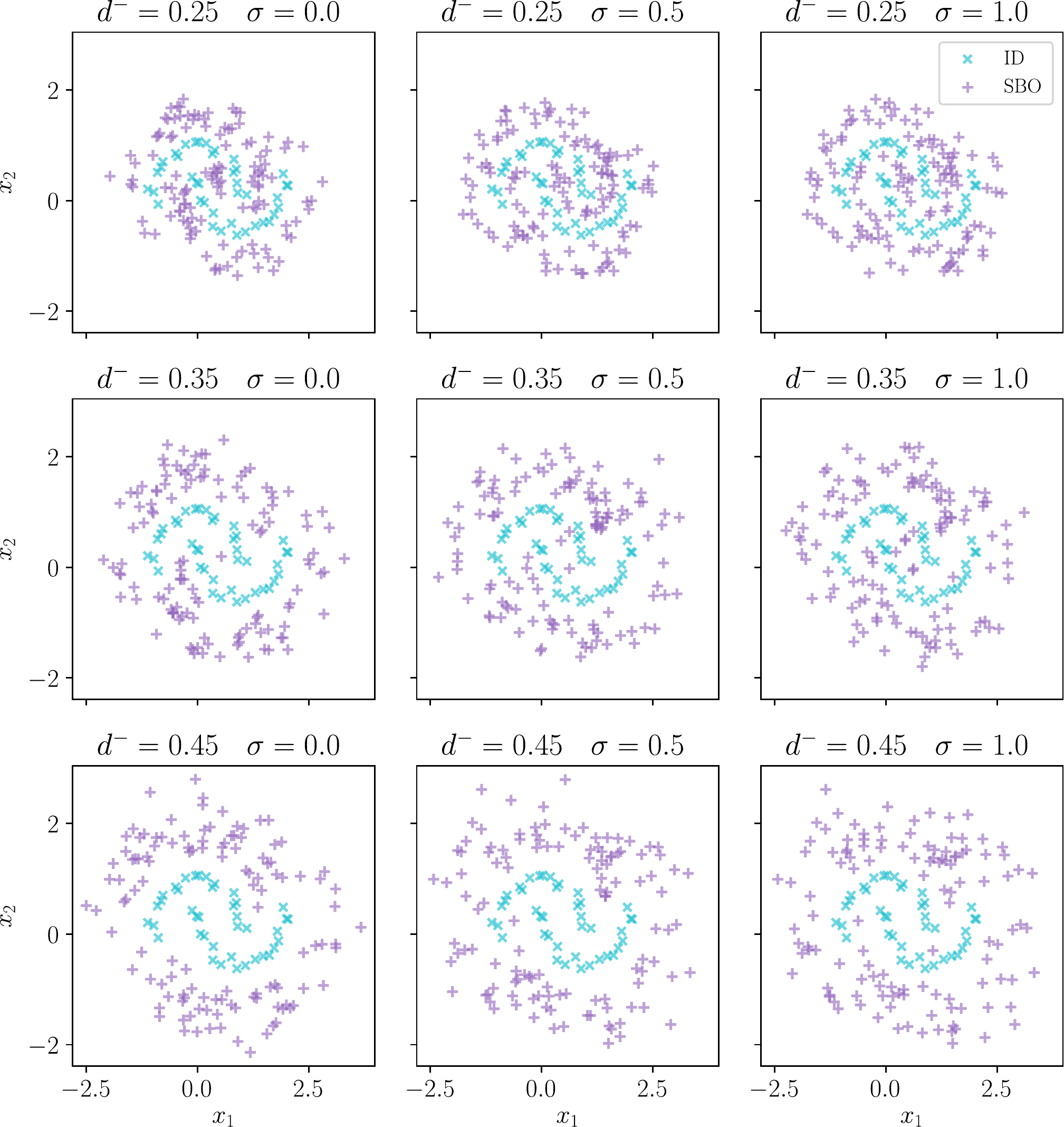
    
    \caption{Exemplary choice of parameters for SBO: ID dataset was sampled using Scikit-learn's function \emph{make\_moons}~\cite{PVG+11} while SBO's parameters are annotated with varying $d^-$ and $\sigma$ while it holds that $d^+=d^-$. The bottom row makes the difference of the softness parameter visible where it regulates whether OOD samples creep in on pockets ($\sigma > 0$) or not ($\sigma = 0$). Data resulting from application of GHO would not account for the pockets between the two shapes~--~whereas SBO's advantage is most prominently noticeable in the first row~--~but would instead create a circular shape as seen in Fig.~\ref{fig:gho}. It is noteworthy that density transfers from ID to OOD originating in its uniform sampling of $\mathbf{y}$ (cf. Alg.~\ref{alg:sbo}).}
    \label{fig:sbo}
\end{figure}

\subsection{Soft Brownian Offset} 

Soft Brownian Offset (SBO) defines an iterative approach to translate a point $\bm{x} \in \bm{X}$ by a most likely distance $d^-$ from a given dataset $\bm{X}$ (cf. Fig.~\ref{fig:sbo}).
It is inspired by Brownian motion~(BM)~\cite{Ein56} and transfers the concept from its one-dimensional origin to $\mathbb{R}^D$ using hyperspheres as a topological basis.
It shares a loose connection with Gaussian processes, since BM has properties of one within the Wiener process~\cite{Par81}
which is also confirmed by Donsker's theorem, a functional extension to the central limit theorem~\cite{Don51,Don52}.

As shown in Algorithm~\ref{alg:sbo}
the approach first uniformly selects a sample from the original dataset and then transforms iteratively until the transformed data point's minimum distance to the dataset $d^* = d_{\min}(\bm{y}, \bm{X})$ transgresses the boundary distance of $d^-$.
Allowing for soft boundaries, the rejection likelihood $\rho \in [0, 1]$ decides on whether to stop early.
It is defined as

\begin{equation}
\rho(d^*, d^-, \sigma) = \left(1 + \exp\left(\frac{d^* + d^-}{\sigma d^- \kappa}\right)\right)^{-1}
\end{equation}

and is falling sigmoidically with an increasing minimum distance $d^*$ and most likely distance $d^-$.
To accomodate for a likelihood $\rho$ sufficiently close to 1 for $d^* = 0$, $\kappa = 7$ is a proposed common choice while $\sigma \in [0, 1]$ defines the boundary softness.

\begin{algorithm}[t]
\caption{Brownian Offset}\label{alg:sbo}
    \KwData{ID samples $\bm{X}$ with individual samples $\bm{x} \in \mathbb{R}^D$ and $|\bm{X}| = N$, Most likely distance $d^-$, Offset distance $d^+$, Boundary softness $\sigma$}
    \KwResult{OOD samples $\bm{Y}$ with individual samples $\bm{y} \in \mathbb{R}^D$ and $|\bm{Y}| = M$}
\For{$i \in \{1, \dots, M\}$}{
    $\bm{y} \leftarrow $ Uniform$(\bm{X})$\;
    $d^* \leftarrow 0$\;
    \While{$d^* < d^-$}{
        $\bm{y} \leftarrow \bm{y} + d^+ \gamma\left(1, 1\right)$\;
        $d^* \leftarrow d_{\text{min}}(\bm{y}, \bm{X})$\;
        $u \leftarrow$ Uniform$([0, 1])$\;
        \If{$u < \rho(d^*, d^-, \sigma) $}{
            \textbf{break}
        }
    }
    $\bm{Y}_i \leftarrow \bm{y}$\;
}
\Return $\bm{Y}$
\end{algorithm}

\subsection{Hard Brownian Offset} 

Hard Brownian Offset (HBO) can be recovered as a special case from Algorithm~\ref{alg:sbo} using $\sigma = 0$.
It then converts the most likely distance $d^-$ to a guaranteed minimum distance through neglection of probabilistic break statements given in regular SBO.
Disabling boundary softness leads to a strict demarcation which can be a desired feature depending on the application.

\section{Experiments} \label{sec:experiments}

Our experiments are three-fold: First, we demonstrate the approaches' capabilities to sample OOD time series. Second, we present a quantitative evaluation regarding the potential improvement for OOD detection on image data. We conclude our experimental evaluation with a case study on the synthetic generation of OOD samples for cyclist trajectories.

\subsection{Time series}

\begin{figure}[tb]
    \centering

    \def\svgwidth{\columnwidth}
    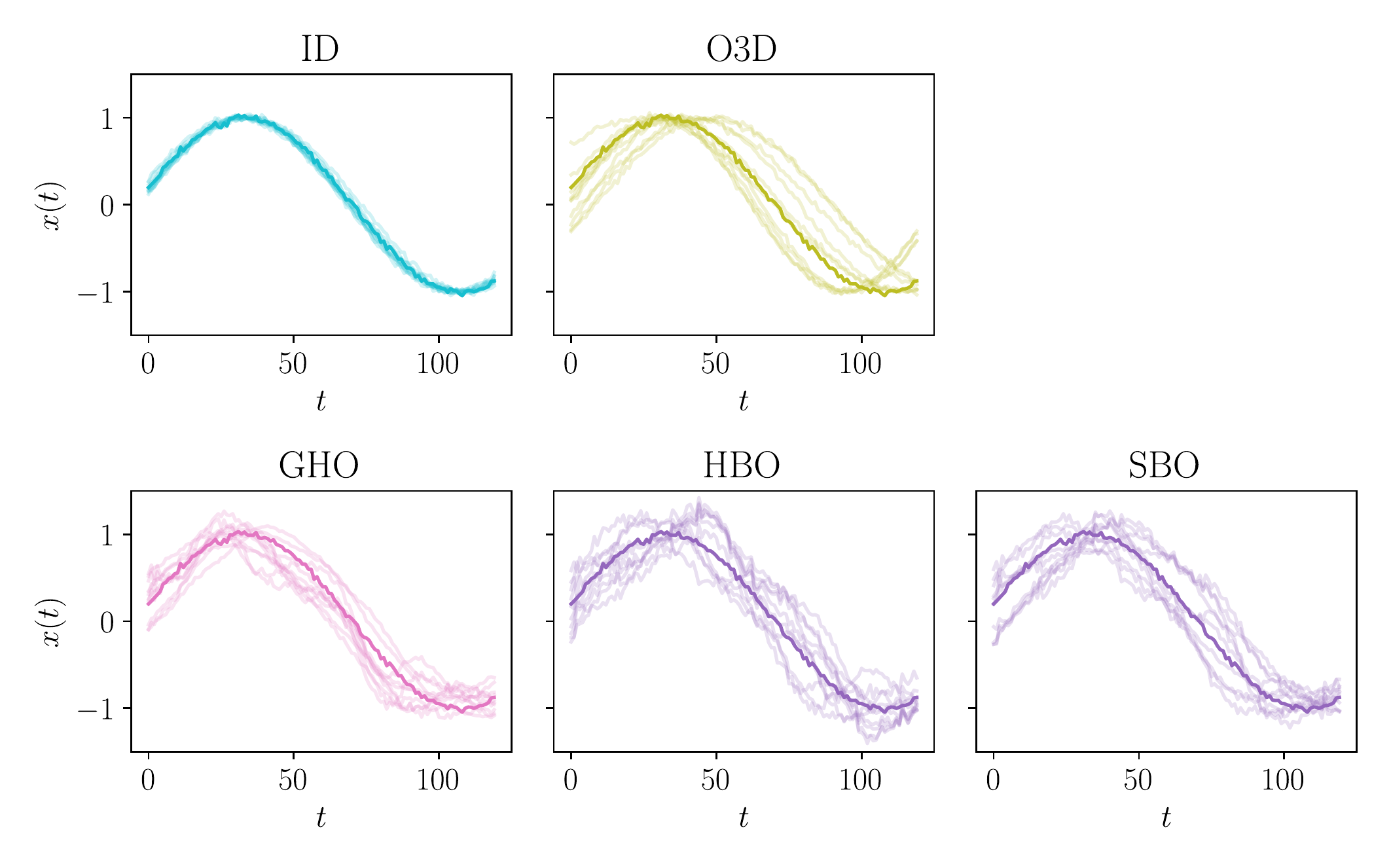
    
    \caption{Archetype class of ID dataset (bold) and selected samples closest to it from selected dataset (transparent) according to DTW distance. O3D's inavailability in real settings stemming from its synthetic nature has to be emphasized.}
\label{fig:archetypes}
\end{figure}

For time series, a synthetic ID baseline dataset of sine waves is created, where a single wave is given by 
\begin{align}
    \text{gt}(t) = \sin(2\pi\cdot f \cdot t) + \epsilon
\end{align}
with frequency $f \sim \mathcal{N}(0, 35)$, noise $\epsilon \sim \mathcal{N}(0, \num{1e-1})$, and time $t \in [0, \ldots, 125]$. The ID training dataset comprises $2000$ time series sampled from this model (cf.\ Fig.~\ref{fig:archetypes}) and is denoted as $\bm{X}_{id}$. All labels are set to $\bm{y}_{id} = 0$ (i.e., ID samples).
Not only does this have the advantage of being hypothetically reducible to a single dimension in $\bm{Z}_{id}$ (cf.\ the spectral decomposition of the input signal) but also of knowing the underlying distribution for $\bm{Z}_{id}$ and inductively $\bm{X}_{id}$, which is a desired property to form a comparable OOD baseline.

Having created an ID dataset, $\bm{X}_{id}$ is then used to train a VAE matching the identity while minimizing reconstruction loss.
The actual architectures for the VAE's encoder and decoder are chosen to be
\begin{align}
\text{Encoder} &\equiv L_{64} \times L_{48} \times L_{32} \times L_{20},\\
\text{Decoder} &\equiv L_{20} \times L_{32} \times L_{48} \times L_{120},
\end{align}
where $L$ indicates linear layers with number of neurons denoted in the index. We use rectified linear units as activation functions.

Having fully trained the VAE, the encoder is used to transform $\bm{X}_{id}$ to generate its learned latent representation $\hat{\bm{Z}}^{\text{vae}}$.
These serve as the basis for the application of the proposed approaches of GHO, HBO and SBO. In particular, we use these methods to generate OOD samples in the latent representation denoted as $\hat{\bm{Z}}^{\text{vae}}_{\text{GHO}}$, $\hat{\bm{Z}}^{\text{vae}}_{\text{HBO}}$ and $\hat{\bm{Z}}^{\text{vae}}_{\text{SBO}}$.

Having created these OOD samples, we use the VAE's decoder to reconstruct OOD samples in the input space. These are referred to  $\bm{X}_{\text{GHO}}$, $\bm{X}_{\text{HBO}}$ and $\bm{X}_{\text{SBO}}$, respectively.
All of these datasets have labels of $\bm{y}_{ood} = 1$, i.e., they contain only OOD samples.

\begin{table}[htb]
    \caption{Hyperparameter settings of the OOD sampling methods.}\label{tbl:hyper_params}
    \centering
        \begin{tabular}{l|c}
            Method & Parameterization  \\\hline
            GHO & $\mu = 9$ and $\sigma = 0.8$  \\
            HBO & $d^* = d^+ = 2$ and $\sigma_{\text{HBO}} = 0$ \\
            SBO & $d^* = d^+ = 2$ and $\sigma_{\text{SBO}} = 1$  \\
        \end{tabular}
\end{table}
The hyperparameter settings are depicted in Tab.~\ref{tbl:hyper_params}. We generate $\num{2000}$ samples using each method.
Figure~\ref{fig:archetypes} shows a selection of produced results.
%

To further validate our OOD sampling methods, we construct an ``optimal'' OOD baseline strategy using the knowledge about the data generating process, i.e., spectral composition underlying the generation of sine curves.
The creation of this synthetic OOD baseline dataset $\bm{X}_\text{O3D}$ follows a process similar to $\bm{X}_{\text{ID}}$, with the difference that the we only sample from the tails of the data generating distribution. 

Next, we train an OOD detector in a supervised fashion. 
The baseline dataset for training an OOD detector is given by the union of $\bm{X}_{\text{ID}}$ and $\bm{X}_\text{O3D}$. We consider the OOD samples created by the baseline as ``ground truth''. 
We use our different datasets for training the OOD detector, i.e., the artificially created OOD sets $\bm{X}_{\text{GHO}}$, $\bm{X}_{\text{HBO}}$ and $\bm{X}_{\text{SBO}}$ and compare it against the baseline OOD set $\bm{X}_\text{O3D}$. These datasets are combined with the $\bm{X}_{\text{ID}}$ to obtain the respective dataset for training the OOD detector.
Training and test split ratio was chosen to be 9:1.
Table \ref{tbl:ds} shows how the selection of datasets influences model metrics in this setting by comparing training data sets with their respectively trained discriminator's $F_1$ score. $\hat{F_1}$ is given by 
\begin{align}
    \hat{F_1}(\bm{X}) &= \frac{F_1(\bm{X})}{F_1(\bm{X}_\text{ID} \cup \bm{X}_\text{O3D})} - 1
\end{align}
and describe the change of the score relative to the baseline dataset. We show the mean of $100$ individual trial runs.
We see that the OOD detector trained with our proposed OOD sampling methods performs comparable.


%
%

\begin{table}[htb]
    \caption{Influence of data set selection on model metrics}
    \label{tbl:ds}
    \centering
        \begin{tabular}{lrr}
            $\bm{X}^{\text{train}} \in$                          & $\hat{F_1}$  \\\hline
            $\bm{X}_{\text{ID}} \cup \bm{X}_{\text{O3D}}$ & \bf{0.00}          \\
            $\bm{X}_{\text{ID}} \cup \bm{X}_{\text{GHO}}$ & -0.02                \\
            $\bm{X}_{\text{ID}} \cup \bm{X}_{\text{HBO}}$ & -0.01                \\
            $\bm{X}_{\text{ID}} \cup \bm{X}_{\text{SBO}}$ & -0.01                
        \end{tabular}
\end{table}

Table~\ref{tbl:wasserstein_dtw}
describes the Wasserstein distance between two datasets using Dynamic Time Warping (DTW)~\cite{Mul07} to measure distance between individual samples. This provides an intuition about the similarities of the generated OOD samples with respect to the methods used, the baseline, and the ID dataset.
The distance matrix $\hat{\Delta}$ between datasets is defined row and column-wise as
\begin{equation}
    W_{d}^{\text{(norm)}}(P_i, P_j) = W_{d}(P_i, P_j) / \max W_{d}(P_x, P_y)
\end{equation}
with $i, j, x, y \in \{ \text{ID}, \text{O3D}, \text{GHO}, \text{SBO}, \text{HBO} \}$.\\
$P_i$ and $P_j$ are shown in the columns and rows respectively, where they are indicated by their corresponding dataset.
Because of metrics' symmetry, only a triangular excerpt is shown.


\begin{table}[tbh]
    \caption{Normalized DTW Wasserstein distance between datasets}\label{tbl:wasserstein_dtw}
    \centering
        \begin{tabular}{c|ccccc}
            $W_d^{(\text{norm})}$  & $\bm{X}_{\text{ID}}$  & $\bm{X}_{\text{O3D}}$ & $\bm{X}_{\text{GHO}}$ & $\bm{X}_{\text{SBO}}$ & $\bm{X}_{\text{HBO}}$ \\
            \hline
            $\bm{X}_{\text{ID}}$    & 0     &  0.375 & 0.818 & \bf{0.835} & 0.761  \\
            $\bm{X}_{\text{O3D}}$   &  &  0     & 0.976 & \bf{1.000} & 0.948  \\
            $\bm{X}_{\text{GHO}}$   &  &   & 0     & \bf{0.230} & 0.204 \\
            $\bm{X}_{\text{SBO}}$   &  &   &  & 0     & 0.026 \\
            $\bm{X}_{\text{HBO}}$   &   &    &   &   & 0 \\
        \end{tabular}
\end{table}

Interpreting these results in light of the definition of OOD detection given earlier that we aim to  separate  samples  originating  of  the ID from those of the ODD. In this context, it can be said that SBO seems most promising in OOD generation as it has the highest Wasserstein distance to not only all other generated datasets but also the original ID dataset. Moreover, we also note, that all presented methods have a higher Wasserstein distance to the ID dataset than the baseline $\bm{X}_{\text{O3D}}$. This hints that the generated samples are further away from the ID than the baseline.


\subsection{Image data}

\begin{figure}[h!]
    \centering

    \def\svgwidth{\columnwidth}
    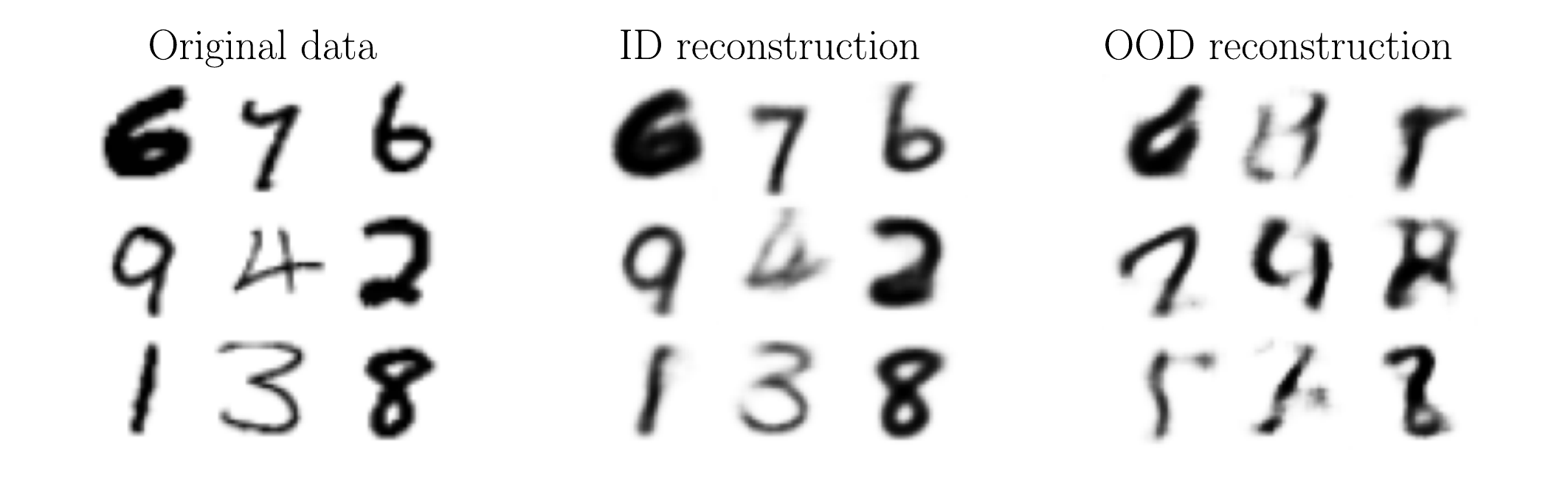
    
    \caption{{\em From left to right}:
        Original images of the MNIST dataset;
        ID reconstructions from a VAE trained on MNIST;
        OOD reconstruction from the same VAE using SBO to transform ID samples within the latent space of the VAE.
        Plots use matching indices which explains partial similarity seen in the OOD reconstructions.
    }
    \label{fig:mnist}
\end{figure}

In this section, we transfer the concept to image data and use the same procedure as described above on the MNIST dataset to generate OOD images depicted in Fig.~\ref{fig:mnist}. Therefore, we train a VAE on the MNIST dataset and apply our OOD detection methods to the learned latent representation. Besides that, we also consider a different (but similar) dataset Not\-MNIST~\cite{notmnist_dataset} (i.e., collection of fonts and glyphs), which we use to evaluate our methods' capability to provide suitable OOD samples to train OOD detectors.

We train separate OOD detection networks using samples from all proposed methods. For OOD detection, we use the method presented by Huseljic et al.~\cite{huseljic2020separation}. As a baseline, we use a OOD detection network trained without any synthetic OOD samples.
All networks use the LeNet architecture~\cite{lecun1998LeNet}.
To evaluate an OOD detectors' ability to detect OOD samples, we evaluate the \textit{Area Under Receiver Operating Characteristic Curve} (AUROC), which is a measure indicating the ability to separate ID and OOD samples. 
We compute it by considering a binary classification problem (MNIST vs.~NotMNIST) where we use the estimated uncertainty from our model and determine whether a sample originates from either ID or OOD regions.
More specifically, we compute the area under the graph that is obtained by plotting the true positive rate against the false-positive rate.
In contrast to other measures, the AUROC is independent regarding the threshold and, therefore, is often used to evaluate methods used for OOD detection~\cite{Malinin2018}.
\begin{table}[h!]
    \caption{Mean Results ($\pm2\sigma$) over five repetitions for OOD detection on Mnist (ID) vs.~NotMNIST (OOD).}\label{tab:mnist_results}
    \centering 
    \begin{tabular}{l|c}
        {} &  AUROC \\\hline
    
        no OOD &      0.9685$\pm$0.0037  \\
        GHO    &      0.9888$\pm$0.0049  \\
        HBO    &      \bf{0.9908}$\pm$\bf{0.0028}  \\
        SBO    &      0.9843$\pm$0.0045  \\
    \end{tabular}
\end{table}

By using our synthetically generated OOD samples, we can see that the OOD detection capability of the neural network is greatly improved. Moreover, we see that HBO (as a special variant of SBO) scores best.


\subsection{Cyclist Trajectories}
\begin{figure}[h!]
    \centerline{\includegraphics[width=\linewidth]{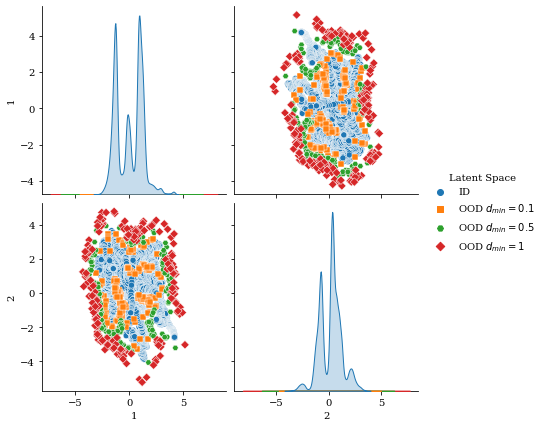}}
    \caption{Two-dimensional latent space representation of trajectories (ID blue dots) using a VAE and OOD samples (OOD colored dots) generated with the SBO method using three different values for the parameter $d_{min}$.}
\label{fig:vae_lat_traj}
\end{figure}

This section presents a case study regarding the application of SBO for cyclist trajectories generation at an experimental research intersection~\cite{GSM12}. The generation of uncommon and still not unrealistic trajectories is an essential building block for safeguarding, i.e., validating the generalization performance, of AI-based trajectory prediction methods in automated vehicles~\cite{Bie21}. Furthermore, increasing the amount of data is essential to optimize the training process, as the number of abnormal instances in the training data is usually sparse.
The cyclists' head positions in the following example are tracked in a two-dimensional coordinate system based on a triangulation technique~\cite{BRZ+17} using two cameras permanently positioned at the intersection. The data consists of 1032 head trajectories, which we split into three different sets for training, evaluation, and testing. The training dataset with 715 trajectories is used to train a VAE to model the distribution of cyclists' movement at the experimental intersection using a low dimensional probabilistic latent representation. The input space of the VAE consists of trajectory segments with a length of 5 seconds, i.e., 250 two-dimensional (x- and y-axis) positions for a sampling rate of \SI{50}{\hertz}. The encoder comprises two hidden layers, i.e., the first hidden layer has 50 neurons, and subsequently, the bottleneck layer describes a two-dimensional latent space. The decoder presents a mirrored structure of the encoder. Moreover, for training, we use the Adam optimizer~\cite{KB15}. As reconstruction loss, we use the mean-squared error of the reconstructed positions.
We depict the two-dimensional latent representation of the VAE in Fig.~\ref{fig:vae_lat_traj}. 

\begin{figure}[h!]
    \centerline{\includegraphics[width=\linewidth]{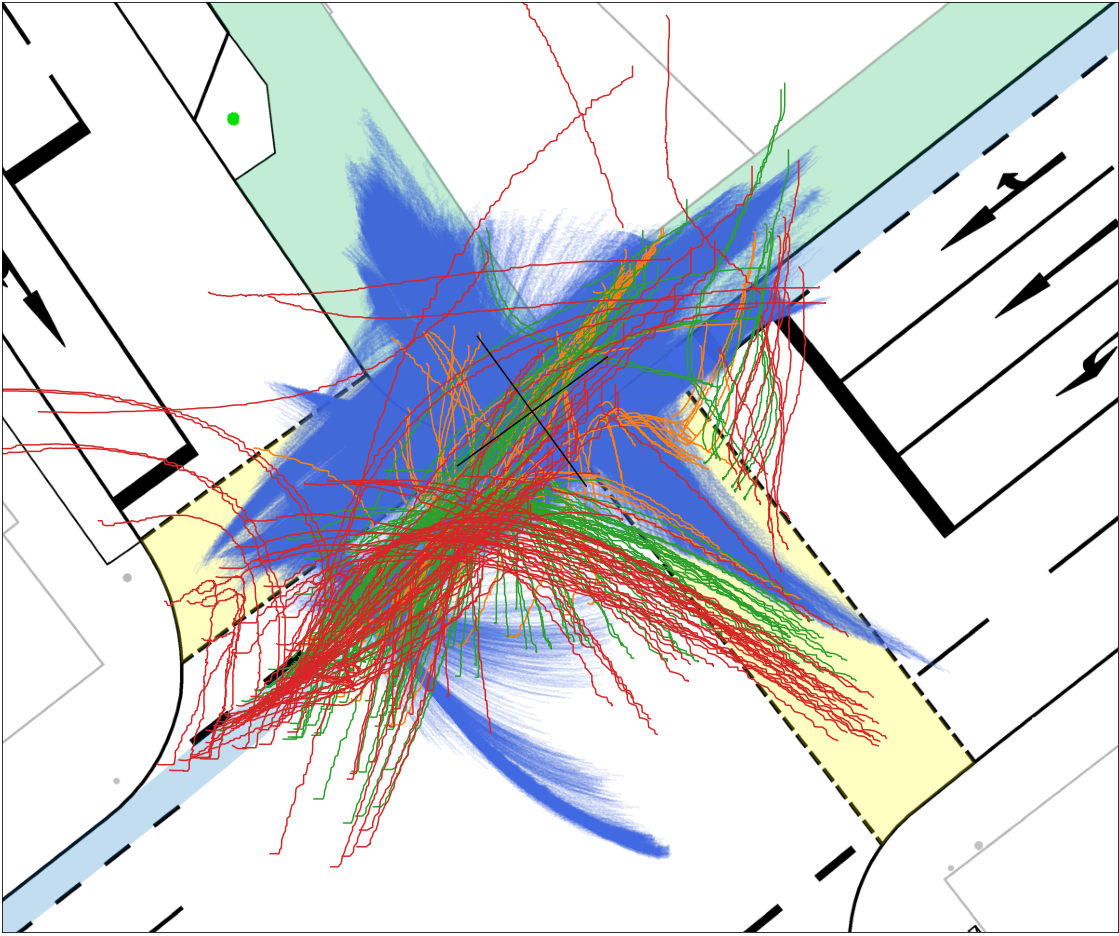}}
    \caption{Top view of the experimental intersection with ID and OOD trajectories. The blue trajectories represent the ID data, and the trajectories in orange, green, and red represent the synthetically generated OOD samples.}
\label{fig:recon_int}
\end{figure}

The generation of OOD instances with our SBO method is performed with an offset distance $d^+=0.1$ and a softness value of $\sigma=0.5$. In this experimental scenario, we focus on showing the influence of the minimal distance parameter $d_{min}$ for generating cyclist trajectories. We investigate for the SBO procedure three different values for the minimal distance $d_{min}$ of $0.1$, $0.5$, and $1$. The test dataset's reconstruction (i.e., ID data) results are presented in Fig.~\ref{fig:recon_int} in the blue region. 

The OOD trajectories generated within three different parameter values for $d_{min}$ are shown in Fig.~\ref{fig:recon_int}. We can appreciate a high correlation between the value of the parameter $d_{min}=1$ and the appearance of the reconstructed red trajectories, as they are abnormal instances which are significantly different than the blue ones. Most likely, these trajectories are more unrealistic trajectories (e.g., cross building). The OOD instances created with the parameter values of $d_{min}=0.1$ and $d_{min}=0.5$ are more realistic. These reconstructions represent vital OOD samples, which are close to the ID representation. 

\section{Conclusion \& Outlook} \label{sec:conclusion}
This article proposed GHO, HBO, and SBO, three strategies for OOD generation. Moreover, we proved the successful application of the OOD sampling methods with synthetic time series data, images, and cyclists' trajectories. We demonstrated that the methods could improve OOD detection and serve as a basis for safeguarding AI-based trajectory prediction methods in the automated driving domain.

The results obtained so far for OOD generation are extremely promising and an important starting point for future research in the field.  
As it turned out in our experiments with time series, one of the main critical points for successful OOD sampling is a proper representation and probabilistic model of the data. With this respect, we aim to investigate more sophisticated representation learning techniques such as SIREN~\cite{SMB+20} or VAE-LSTM~\cite{CSZL19} to model the data. In this context, one focus is the investigation of modeling techniques that better represent similarities between samplings in original and latent space (e.g., geodesic)~\cite{CKFBS+20}. 
Another line of investigation is the generation of cyclist and pedestrian body trajectories (i.e., body poses) using our SBO method. 
Here, we also aim to include additional conditions for the generation of more realistic OOD trajectories. 
For example, we aim to describe pedestrians' motion patterns by a discrete set of motion states (e.g., walking, moving, and turning) and automatically generate appropriate abnormal but still realistic body pose trajectories that can be used to safeguard AI-based prediction models.

\section*{Acknowledgment}
This work results from the project AIMEE (01IS19061) funded by the German Federal Ministry of Education and Research (BMBF), the project KI Data Tooling (19A20001O) funded by the German Federal Ministry for Economic Affairs and Energy (BMWI), and the project DeCoInt$^2$ supported by the German Research Foundation (DFG) within the priority program SPP 1835: “Kooperativ interagierende Automobile”, grant number SI 674/11-2.

\bibliographystyle{ieee_fullname}
\bibliography{bib/references}
\end{document}